# An Unconstrained Symmetric Nonnegative Latent Factor Analysis for Large-scale Undirected Weighted Networks

Zhe Xie, Weiling Li, and Yurong Zhong

**Abstract**—Large-scale undirected weighted networks are usually found in big data-related research fields. It can naturally be quantified as a symmetric high-dimensional and incomplete (SHDI) matrix for implementing big data analysis tasks. A symmetric non-negative latent-factor-analysis (SNL) model is able to efficiently extract latent factors (LFs) from an SHDI matrix. Yet it relies on a constraint-combination training scheme, which makes it lack flexibility. To address this issue, this paper proposes an unconstrained symmetric nonnegative latent-factor-analysis (USNL) model. Its main idea is two-fold: 1) The output LFs are separated from the decision parameters via integrating a nonnegative mapping function into an SNL model; and 2) Stochastic gradient descent (SGD) is adopted for implementing unconstrained model training along with ensuring the output LFs nonnegativity. Empirical studies on four SHDI matrices generated from real big data applications demonstrate that an USNL model achieves higher prediction accuracy of missing data than an SNL model, as well as highly competitive computational efficiency.

——————————— ◆ ———————————

## I. INTRODUCTION

A Large-scale undirected weighted network generated from real big data-related applications contains useful yet implicit knowledge hidden a few relationships such as unfound protein-related information [7, 8, 37], potential community [9, 10], and gene expression prediction [11]. It can be usually described as a symmetric high-dimensional incomplete matrix (SHDI), owing to the following characteristics:
(1) Its entities set is large;
(2) Its most data entries are missing;
(3) It is symmetric;
(4) Its data are commonly nonnegative.
Therefore, how to design an analysis model that fully considers an SHDI matrix's characteristics for extracting such hidden knowledge from the target matrix becomes an increasingly important issue.

Given a full symmetric matrix filled with nonnegative data, a nonnegative matrix factorization (NMF) is proposed [1-3]. It adopts Euclidean distance to build an objective function for mapping latent factors (LFs) into low rank space, and then a non-negative and multiplicative update (NMU) algorithm is designed for solving such an objective function. Liu *et al*. [1] impose the bound constraints on two different LFs, thereby building a novel NMF model that makes it more suitable for clustering applications. Hien *et al*. [2] build an NMF model based on the Kullback–Leibler divergence and design an algorithm for such an NMF model, thereby ensuring global convergence and achieving a fairly good performance. Liang *et al*. [3] propose a novel NMF model which imposes orthogonal constraints on LFs, achieving an improvement in clustering performance. However, they do not consider the third characteristic, i.e., the symmetry of an SHDI matrix when building NMF's objective function. For addressing this issue, a symmetric nonnegative matrix factorization (SNMF) model is proposed, which is similar with NMF and able to analyze such a matrix efficiently for various tasks like community detection [9, 10] and image processing [4, 6]. Belachew *et al*. [4] propose a novel SNMF model that incorporates a sparse constraint on LFs, and then transform it into a non-negative least squares form using a rank-1 approximation for obtaining sparser basic factors to represent image features. Moutier *et al*. [5] propose a SNMF model that ignores diagonal information, thereby making the obtained predictions more accurate and leading to an easier solution of the optimization problem. Li *et al*. [6] convert an SNMF model into an asymmetric NMF model by imposing a penalty term on LFs. Then they adopt an alternate algorithm for solving such a novel model efficiently, as well as demonstrates the proposed model's feasibility. In spite of their effectiveness, they do not consider the first and the second characteristics of an SHDI matrix, i.e., high-dimensionality and incompleteness. In other words, they need to prefill the target SHDI matrix's missing data before their training, which leads to massive time cost and accuracy loss [12, 13].

For addressing above issue, Luo *et al*. [12] propose a symmetric nonnegative latent-factor-analysis (SNL) model whose loss

——————————————————————————————————————————————

- *Corresponding author: Yurong Zhong.*
- *Z. Xie, W. Li, and Y. Zhong are with the School of Computer Science and Technology, Dongguan University of Technology, Dongguan, Guangdong 523808, China (e-mail: gxyz4419@gmail.com, weilinglicq@outlook.com, zhongyurong91@gmail.com).*

function is built via considering all essential characteristics of an SHDI matrix. Then they design a symmetric, single latent factor-dependent, non-negative and multiplicative update (S²LF-NMU) algorithm for solving such a loss function efficiently, thereby obtaining nonnegative latent factors (LFs) from an SHDI matrix with high efficiency in computation and storage.

In spite of an SNL model's efficiency and effectiveness, its adopted S²LF-NMU algorithm manipulates learning rates to cancel the nonnegative terms for ensuring its nonnegative constraint, i.e., it relies on specifically designed learning algorithm. Such a scheme lack of flexibility and scalability in various learning tasks like recommender system [14, 15] and QoS service prediction [16, 17]. For addressing this issue, we propose an unconstrained symmetric nonnegative latent-factor-analysis (USNL) model solved by a general training scheme along with nonnegative constraints fulfilled. The main contributions of this paper include:

(1) An USNL model is proposed. It integrates nonnegative mapping function into an SNL model, which relaxes its nonnegative constraint. Then stochastic gradient descent (SGD) is adopted for solving such an unconstrained loss function, thereby obtaining nonnegative LFs efficiently;
(2) Empirical studies on four SHDI matrices generated from real applications reveal that an USNL model achieves higher prediction accuracy of missing data than an SNL model, as well as highly competitive computational efficiency.

Section II gives the preliminaries. Section III presents an USNL model. Section IV states the experimental results. And finally, Section V draws the conclusions.

## II. PRELIMINARIES

### A. Problem Formulation

The SHDI matrix describes certain relationship among each entity which usually are non-negative. Such a matrix can be described as:

***Definition* 1.** Given a target network $N$, each single entry quantifies some kind of interactions among the matrix $R^{|N|\times|N|}$, which is non-negative. Given known set $\Lambda$ and unknown set $\Gamma$ for $R$, $R$ is SHDI if $|\Lambda|\ll|\Gamma|$.

For extracting potential yet useful information from SHDI, an SNL model is defined as:

***Definition* 2.** Given $R$ and $\Lambda$, an SNL model usually relies on $\Lambda$ to seek for rank-$d$ approximation of $R$, i.e., $\hat{R}=XX^T$ where $X^{|N|\times d} \geq 0$. With commonly-used Euclidean distance [18, 19], the following object function is defined as:

$$\varepsilon = \frac{1}{2}\sum_{r_{i,j}\in\Lambda}\left(r_{i,j} - \sum_{k=1}^{d}x_{i,k}x_{j,k}\right)^2 \quad (1)$$

$$s.t. \quad \forall i,j \in N, k \in \{1,2,\ldots,d\}: x_{i,k} \geq 0, x_{j,k} \geq 0.$$

According to previous researches [20-22], extracting LFs from incomplete matrix is ill-posed, so we adopt $L_2$-norm-based regularization scheme to avoid over-fitting. Then (1) can be rewritten to:

$$\varepsilon = \frac{1}{2}\sum_{r_{i,j}\in\Lambda}\left(\left(r_{i,j} - \sum_{k=1}^{d}x_{i,k}x_{j,k}\right)^2 + \lambda\left(\sum_{k=1}^{d}(x_{i,k})^2 + \sum_{k=1}^{d}(x_{j,k})^2\right)\right) \quad (2)$$

$$s.t. \quad \forall i,j \in N, k \in \{1,2,\ldots,d\}: x_{i,k} \geq 0, x_{j,k} \geq 0.$$

## III. AN USNL MODEL

The latent factor $X$ in object function (2) is used as both the output LFs and the decision parameter, which lacks flexibility. By introducing nonnegative mapping function and $Y$, we achieve relaxation of nonnegative constraint on $X$, and the relationship between $X$ and $Y$ is defined as:

$$\forall i,j \in N, k \in \{1,2,\ldots,d\}:$$
$$x_{i,k} = f(y_{i,k}), x_{j,k} = f(y_{j,k}) \quad (3)$$

By introducing nonnegative mapping function $f$, i.e., fulfilling the condition that $x_{i,k}=f(y_{i,k})\geq 0$ and $x_{j,k}=f(y_{j,k})\geq 0$, an object function (2) can be rewritten in the following form:

$$\varepsilon = \frac{1}{2}\sum_{r_{i,j}\in\Lambda}\left(\left(r_{i,j} - \sum_{k=1}^{d}f(y_{i,k})f(y_{j,k})\right)^2 + \lambda\left(\sum_{k=1}^{d}(f(y_{i,k}))^2 + \sum_{k=1}^{d}(f(y_{j,k}))^2\right)\right) \quad (4)$$

By doing this, we are able to update $Y$ as a decision parameter during the model training while $X$ is kept as the output LFs with nonnegative values.

It is obvious that the choice of mapping function is critical, and its selection is related to the data set, which means that the performance of the same mapping function in different dataset may vary greatly. Note that the mapping functions adopted in this paper will be given in Table I.

TABLE I. Adopted Mapping Function.

| Name | Function | Derivative |
|---|---|---|
| Sigmoid | $f(a) = \dfrac{1}{1+e^{-a}}$ | $f'(a) = f(a)\cdot(1-f(a))$ |
| Absolute | $f(a) = \lvert a \rvert_{abs}$ | $f'(a) = \begin{cases} 1, & \text{if } a > 0 \\ -1, & \text{if } a \leq 0 \end{cases}$ |
| Relu | $f(a) = \begin{cases} a, & \text{if } a > 0 \\ 0, & \text{if } a \leq 0 \end{cases}$ | $f'(a) = \begin{cases} 1, & \text{if } a > 0 \\ 0, & \text{if } a \leq 0 \end{cases}$ |

According to previous research [23-25], an SGD algorithm has good performance on high-dimensional and complete matrix and is easy to be implemented. Hence, according to the inverse direction of the gradient of each training instance, we minimize (4) with SGD, thereby obtaining the following rules:

$$\begin{cases} y_{i,k} \leftarrow y_{i,k} - \eta \dfrac{\partial \varepsilon_{i,j}}{\partial f(y_{i,k})} f'(y_{i,k}), \\ y_{j,k} \leftarrow y_{j,k} - \eta \dfrac{\partial \varepsilon_{i,j}}{\partial f(y_{j,k})} f'(y_{j,k}), \end{cases} \qquad (5)$$

where $\varepsilon_{i,j}$ denotes the instant error on the training instance, and $\eta$ denotes learning rate of gradient descent algorithm. With (4) and (5), we get the following rules:

$$\begin{cases} y_{i,k} \leftarrow y_{i,k} + \eta f'(y_{i,k})\left(f(y_{j,k}) e_{i,j} - \lambda f(y_{i,k})\right), \\ y_{j,k} \leftarrow y_{j,k} + \eta f'(y_{j,k})\left(f(y_{i,k}) e_{i,j} - \lambda f(y_{j,k})\right), \end{cases} \qquad (6)$$

where $e_{i,j} = r_{i,j} - \sum_{k=1}^{d} f(y_{i,k}) f(y_{j,k})$. From (6), the update rules for $y_{i,k}$ and $y_{j,k}$ are the same. So we obtain the following concise form:

$$y_{i,k} \leftarrow y_{i,k} + \eta f'(y_{i,k})\left(f(y_{j,k}) e_{i,j} - \lambda f(y_{i,k})\right). \qquad (7)$$

## IV. EXPERIMENTAL RESULTS AND ANALYSIS

### A. General Settings

**Evaluation Protocol.** For real applications, it is very important to decompose an SHDI matrix into LFs for predicting its missing data, since there exists a strong desire to discover connections between potentially involved entities [12, 26]. Therefore, we adopt it as an evaluation protocol to verify the performance of all related models.

**Evaluation Metrics.** The accuracy of a test model for missing data predictions can be measured by the root mean square error (RMSE) [27-30]:

$$RMSE = \sqrt{\left(\sum_{r_{i,j} \in \Gamma} (r_{i,j} - \hat{r}_{i,j})^2\right) \bigg/ |\Gamma|},$$

where $\Gamma$ denotes the validation set and is disjoint with the training set $\Lambda$. Note that low RMSE represents high prediction accuracy for missing data in $\Gamma$. All experiments are implemented on a tablet with a 3.6-GHz i3 CPU and 16-GB RAM, and implemented in JAVA SE18.

TABLE II. Details of Adopted Datasets.

| No. | $|\Lambda|$ | $|N|$ | Density |
|---|---|---|---|
| D1 | 1021786 | 4181 | 5.85% |
| D2 | 1182124 | 5194 | 4.38% |
| D3 | 2795876 | 2573 | 42.23% |
| D4 | 95188 | 16726 | 0.03% |

TABLE III. Details of Tested Models.

| No. | Name | Description |
|---|---|---|
| M1 | $\beta$-SNMF | An SNMF model solved by an improved NMU-based algorithm [36]. |
| M2 | SNL | An SNL model solved by an S²LF-NMU algorithm [12]. |
| M3 | USNL-SIG | An USNL Model with Sigmoid mapping function solved by an SGD algorithm. |
| M4 | USNL-REL | An USNL Model with Relu mapping function solved by an SGD algorithm. |
| M5 | USNL-ABS | An USNL Model with Absolute mapping function solved by an SGD algorithm. |

**Datasets.** Our experiments adopt four SHDI matrices, and their details are shown in Table II. D1, D2 and D3 describe few known interactions among a large number of proteins in Bacillus Subtilis, Naumovozyma castellii, and Acetobacter aceti 1023, respectively. Note that D1-3 are from STRING database [31]. D4 records few interactions in a collaboration network of preprints at www.arxiv.org, which is from the University of Florida sparse matrix collection [32]. In all experiments on each dataset, we randomly split its known set of entries Λ into ten disjoint subsets for tenfold cross-validation. We adopt seven subsets as training set, one subset as validation set, and the remaining two subsets as test subsets. This process is repeated ten times sequentially.

**Compared Models.** Our experiments involve five models, where, M1-2 is state-of-the-art model in analyzing an SHDI matrix and M3-5 are the proposed USNL models with different mapping functions. Their details are presented in Table III. To achieve its objective results, we used the following settings:
(1) LF Dimension $d$ is set at 20;
(2) For each model on each data set, the results generated from 20 different random initial values are recorded to calculate the average RMSE and convergence time for eliminating the effect of initial assumptions [33-35].
(3) The training process of the test model is terminated when: 1) the iteration count reaches a preset threshold, which is 1000; 2) The difference between the two consecutive iterations of the generated RMSE is less than $10^{-5}$.

*B. Comparison against State-of-the-art Models*

Table IV-VI summarize RMSE, converging iteration count and total time cost of M1-5, respectively. From these results, we have the following findings:
(1) **An USNL model achieves better accuracy than M1 on D1-4.** For example, from Table IV, RMSE of M1-4 on D1 is 0.1543, 0.1327, 0.1303, 0.1325, which is about 15.88%, 2.19%, 0.38%, 2.04% lower than M5's 0.1298, respectively. Similar results can be found on D2-4. From above, we clearly see that integrating different nonnegative mapping functions does not reduce prediction accuracy of missing data from an SHDI matrix.
(2) **Using a suitable mapping function makes the USNL model converge fastest.** As shown in the Table V, USNL models converge with the least iteration count on D1-4. Note that M3 converges fastest on D4, M4 wins on D1 and D3, and M5 wins on D2. Hence, converging iteration count in analyzing on an SHDI matrix is mapping function-dependent.
(3) **Computational efficiency of an USNL model is highly competitive.** From Table VI, among an USNL model with three nonnegative mapping functions, i.e., Sigmoid, Relu and Absolute, M5 consumes the least total time cost on D1-4. Moreover, M5 obtain the least total time cost on D4, and M2 wins on the D1-3.

TABLE IV. RMSE of M1-5 on D1-4.

| No. | M1 | M2 | M3 | M4 | M5 |
|---|---|---|---|---|---|
| D1 | 0.1543±3.2E-4 | 0.1327±2.9E-4 | 0.1303±4.5E-4 | 0.1325±6.3E-5 | **0.1298±6.2E-5** |
| D2 | 0.1806±4.1E-4 | 0.1415±3.8E-5 | 0.1416±2.2E-4 | 0.1410±1.8E-4 | **0.1400±1.5E-4** |
| D3 | 0.1003±0.8E-4 | 0.0926±3.2E-5 | 0.0918±1.3E-4 | 0.0919±5.4E-5 | **0.0916±3.6E-5** |
| D4 | Failure | 0.8048±1.2E-4 | 0.9196±2.1E-2 | **0.7999±2.8E-3** | 0.8003±1.9E-4 |

TABLE V. RMSE of M1-5 on D1-4.

| No. | M1 | M2 | M3 | M4 | M5 |
|---|---|---|---|---|---|
| D1 | 688±22.35 | 350±2.32 | 215±6.53 | **94±1.92** | 157±7.05 |
| D2 | 959±29.28 | 113±1.68 | 366±18.36 | 175±2.86 | **97±5.59** |
| D3 | 290±4.15 | 611±5.31 | 134±9.76 | **43±1.14** | 50±2.05 |
| D4 | Failure | 186±2.47 | **181±23.41** | 448±52.91 | 314±82.49 |

TABLE VI. Total Time Cost of M1-5 on D1-4 (Seconds).

| No. | M1 | M2 | M3 | M4 | M5 |
|---|---|---|---|---|---|
| D1 | 1159±109.90 | **101±1.38** | 696±11.77 | 170±7.66 | 106±7.77 |
| D2 | 1793±164.29 | **123±6.36** | 792±9.95 | 232±6.26 | 138±6.26 |
| D3 | 435±37.99 | **241±4.08** | 1835±12.02 | 406±18.26 | 293±11.45 |
| D4 | Failure | 20±0.65 | 78±2.82 | 19±1.31 | **16±0.48** |

V. CONCLUSIONS

This paper proposes an USNL model for better effectively predicting missing data from an SHDI matrix than an SNL model. Its main idea is to separate the output LFs from the decision parameter via integrating a nonnegative mapping function into an SNL model, thereby transforming a nonnegativity-constrained problem into an unconstrained problem on an SHDI matrix. Then an SGD algorithm is adopted for solving such an unconstrained problem efficiently. Empirical studies on four SHDI matrices show that the proposed model is able to precisely represent the symmetry of an SHDI matrix, and achieve better accuracy gain than an SNL model with highly competitive computational efficiency.

However, there exists unnecessary information loss in extracting nonnegative LFs from an SHDI matrix in an USNL model, due to its strong symmetry assumption based on one unique latent factor matrix. Hence, we will investigate how to relax such an assumption in an USNL model for seeking higher precision representation to an SHDI matrix.